\documentclass[letterpaper, 10 pt, conference]{ieeeconf}  

\IEEEoverridecommandlockouts                              

\usepackage{amsmath}
\usepackage{amssymb}
\usepackage{algorithm}
\usepackage{algpseudocode}
\usepackage{caption}
\usepackage{xcolor}
\usepackage{float}
\usepackage{lipsum}
\usepackage{cite}
\usepackage{hyperref}
\usepackage{booktabs}
\floatstyle{ruled}
\restylefloat{algorithm}
\usepackage{graphicx}

\usepackage{algorithm}
\usepackage{algpseudocode}
\usepackage{amsmath}
\usepackage{tikz}
\usetikzlibrary{arrows.meta,calc,positioning}
\usepackage{comment}

\definecolor{algcomment}{RGB}{30,150,30}

\newcommand{\vect}[1]{\boldsymbol{#1}}
\newcommand{\norm}[1]{\left\lVert #1\right\rVert}

\title{\LARGE \bf
RoKON: Efficient and Adaptive Data-Enabled Predictive Control \\ for  Robotic Manipulation
}

\author{Akshun Sharma$^{1}$, Zhongqi Wei$^{2}$, Buqing Ou$^{2}$, Frederike D\"{u}mbgen$^{2}$
\thanks{$^{1}$Maryland Applied Graduate Engineering, University of Maryland, College Park, MD 20742, USA {\tt\small asharm41@umd.edu}}
\thanks{$^{2}$Department of Mechanical Engineering, Carnegie Mellon University, Pittsburgh, PA 15213, USA {\tt\small \{zhongqi2, buqingo, fdumbgen\}@andrew.cmu.edu}.}
}

\begin{document}

\maketitle
\thispagestyle{empty}
\pagestyle{empty}

\begin{abstract}

Data-enabled predictive control has recently emerged as a promising control paradigm, enabling a transparent optimal-control formulation without the need to learn an explicit dynamics model. 
This paper presents a number of extensions paving the path towards  its use in real-time robotics tasks. 
Following recent work, we utilize a nonlinear predictor based on Gaussian kernels to represent the relationship between input and output data. 
We propose a damped Gauss--Newton method that enables real-time control by efficiently exploiting the least-squares structure of the control problem. In addition, we introduce a hybrid buffer management strategy that incorporates informative online data while retaining relevant offline data, allowing the predictor to account for changes in the dynamics. We demonstrate the tracking performance and computational suitability of the
proposed method in MuJoCo using a Franka Research~3 arm pushing a block across surface, without prior knowledge of its changing friction values.

\end{abstract}

\section{INTRODUCTION}

Robotic systems increasingly operate in settings where an accurate model of the environment is difficult to obtain. Contact, friction, and actuator dynamics can introduce behaviors that are expensive to model and identify, particularly when a robot interacts with previously unseen objects or surfaces~\cite{billard2019trends}. Data-driven control is a control paradigm that either identifies a system model from measured data and subsequently uses it to determine control actions (indirect data-driven control)~\cite{hou2013model,hewing2020learning}, or directly constructs control actions from measured input--output data without identifying an explicit model (direct data-driven control)~\cite{coulson2019deepc}.

Large-scale robot-learning efforts have demonstrated the benefits of
aggregating diverse datasets~\cite{openx2024}. Collecting such datasets on
physical robots, however, requires substantial time, supervision, and hardware access, while exploratory actions may result in task failures or hardware damage~\cite{muratore2022randomized,kroemer2021review}. In such cases, the controller must extract useful dynamical information from a comparatively small number of input--output measurements. Prior work has demonstrated that data-driven control can achieve accurate closed-loop performance using only a small number of training samples~\cite{bauza2018dataefficient, persson2026adaptive, alsalti2024efficient}. 

A prominent framework for direct data-driven control is Data-enabled Predictive Control (DeePC) \cite{coulson2019deepc}. DeePC builds on Willems' fundamental lemma, which states that, for a controllable linear time-invariant (LTI) system driven by a sufficiently persistently exciting input, the columns of a Hankel matrix constructed from one measured trajectory span all finite-length trajectories of the system \cite{willems2005fundamental}. DeePC embeds this data-based trajectory
representation into a constrained finite-horizon optimization problem and solves it in a receding-horizon manner without explicitly identifying a state-space model.

Despite extensive developments in the control community, DeePC has only recently begun to receive attention for robotic systems. Existing studies have demonstrated that DeePC can achieve effective closed-loop
control in several robotic settings, including trajectory tracking in aerial
robotics~\cite{coulson2022distributionally}, soft-robot
control~\cite{wang2026velocity}, and simulated robotic-arm tasks~\cite{ouyang2025direct}. Although these studies demonstrate its practical potential beyond strictly linear systems, the trajectory parameterization underlying standard DeePC is exact only for controllable linear time-invariant systems. Its application to nonlinear robotic systems therefore commonly relies on locally linear representations constructed around the current operating condition~\cite{erust2026safety}, or on lifting the measured variables into a feature space in which the nonlinear dynamics admit an approximately linear representation~\cite{bruder2021koopman}.

A more recent extension, Robust and Kernelized DeePC
(RoKDeePC)~\cite{huang2024rokdeepc}, addresses nonlinear behavior
differently by representing the multistep input--output mapping in a
reproducing kernel Hilbert space (RKHS)~\cite{lian2021nonlinear}. Kernel
evaluations measure similarity between recorded trajectories and provide an
implicit representation of the nonlinear multistep predictor without
constructing an explicit nonlinear state-space model or selecting a
finite-dimensional set of lifting functions. RoKDeePC embeds this kernelized predictor directly into a constrained receding-horizon control problem. However, the resulting optimization is nonconvex and requires more computational effort compared to DeePC. Furthermore, the original formulations of DeePC and RoKDeePC use a fixed offline dataset, and its computational cost increases as that dataset (buffer) grows. To address these limitations, we introduce RoKON, an adaptive kernelized
DeePC framework designed for real-time control under changing dynamics.
RoKON retains the nonlinear dependence of the Gaussian kernel on the future inputs and updates a fixed-capacity hybrid data buffer using informative online measurements. We evaluate RoKON on a robotic pushing task across surfaces with varying friction. 
The contributions of this work are threefold:
\begin{itemize}
    \item We propose a damped Gauss--Newton method to solve the nonconvex RoKDeePC problem efficiently. Compared with the original projected-gradient method used in RoKDeePC, our method exploits the least-squares structure of the control problem, making it significantly faster.

    \item We introduce a hybrid data buffer strategy in which a small offline recorded dataset initializes a fixed-capacity buffer and online trajectories progressively replace a subset of prior data. The buffer admits informative samples only, allowing the predictor to adapt to changing friction surfaces while maintaining bounded memory and computation.

    \item We validate RoKON in MuJoCo simulation on a robotic pushing task under varying friction conditions, and show favorable performance compared with MPC and state-of-the-art data-enabled predictive control baselines. 
\end{itemize}
The remainder of this paper is organized as follows. Section~\ref{Sec:Background} reviews related work on data-driven control, DeePC, and RoKDeePC. Section~\ref{Section: Method} presents our proposed method. Section~\ref{Section: Experiments} describes the setup, problem statement and presents the experiments conducted and discusses the results. Finally, Section~\ref{Sec: Conclusion} summarizes the main findings and discusses directions for future work.

\section{Background}
\label{Sec:Background}

\subsection{Data Driven Control and DeePC}

Data-driven control methods can be broadly divided into two categories:
indirect methods, which identify a model from measured data and subsequently
use it to formulate a control problem, and direct methods, which compute
control actions directly from input--output data. Data-enabled predictive control (DeePC) belongs to the latter class of control algorithms.~\cite{coulson2019deepc}

Consider $T$ offline input--output samples
$\{(u_k^{\mathrm d},y_k^{\mathrm d})\}_{k=0}^{T-1}$, where
$u_k^{\mathrm d}\in\mathbb R^m$ and
$y_k^{\mathrm d}\in\mathbb R^p$ are the measured input and output,
respectively, $k$ is the sample index, and the superscript $\mathrm d$
denotes offline data. The recorded input--output trajectory can be arranged into the block Hankel
matrix of depth $L$, where \(T \geq L > \ell\), with \(\ell\) denoting the lag of the system and each column represents a contiguous input--output trajectory.
\begin{equation}
\mathcal H_L(\vect u^{\mathrm d},\vect y^{\mathrm d})
=
\begin{bmatrix}
u_0^{\mathrm d}     & u_1^{\mathrm d}     & \cdots & u_{T-L}^{\mathrm d}\\
u_1^{\mathrm d}     & u_2^{\mathrm d}     & \cdots & u_{T-L+1}^{\mathrm d}\\
\vdots               & \vdots               & \ddots & \vdots\\
u_{L-1}^{\mathrm d} & u_L^{\mathrm d}     & \cdots & u_{T-1}^{\mathrm d}\\[1mm]
y_0^{\mathrm d}     & y_1^{\mathrm d}     & \cdots & y_{T-L}^{\mathrm d}\\
y_1^{\mathrm d}     & y_2^{\mathrm d}     & \cdots & y_{T-L+1}^{\mathrm d}\\
\vdots               & \vdots               & \ddots & \vdots\\
y_{L-1}^{\mathrm d} & y_L^{\mathrm d}     & \cdots & y_{T-1}^{\mathrm d}
\end{bmatrix}
=
\begin{bmatrix}
\mathcal H_L(\vect u^{\mathrm d})\\
\mathcal H_L(\vect y^{\mathrm d})
\end{bmatrix}.
\label{eq:hankel_definition}
\end{equation}

For the collected trajectories to represent the true system behavior, the
system must be persistently excited by applying rich input data that capture its different behaviors. An input trajectory
$\vect u^{\mathrm d}$, with $u_k^{\mathrm d}\in\mathbb R^m$, is persistently
exciting of order $q$ if
$\operatorname{rank}\bigl(\mathcal H_q(\vect u^{\mathrm d})\bigr)=mq$. This rank condition ensures that the collected data contain sufficient
input variation to represent all admissible input trajectories.

For DeePC, the trajectory length is divided into immediate past $T_{\mathrm{ini}}$ and a future prediction horizon $N$, such that
$L=T_{\mathrm{ini}}+N$. The corresponding input and output Hankel matrices
are partitioned into past and future blocks as
\begin{equation}
\begin{bmatrix}
U_{\mathrm P}\\
U_{\mathrm F}
\end{bmatrix}
=
\mathcal H_{T_{\mathrm{ini}}+N}(\vect{u}^{\mathrm d}),
\qquad
\begin{bmatrix}
Y_{\mathrm P}\\
Y_{\mathrm F}
\end{bmatrix}
=
\mathcal H_{T_{\mathrm{ini}}+N}(\vect{y}^{\mathrm d}),
\label{eq:hankel_partition}
\end{equation}
where $U_{\mathrm P}\in\mathbb{R}^{mT_{\mathrm{ini}}\times H_c}$, $U_{\mathrm F}\in\mathbb{R}^{mN\times H_c}$, ${Y}_{\mathrm P}\in\mathbb{R}^{pT_{\mathrm{ini}}\times H_c}$, ${Y}_{\mathrm F}\in\mathbb{R}^{pN\times H_c}$, and $H_c=T-T_{\mathrm{ini}}-N+1$.

For a controllable LTI system of order $n$, an input that is
persistently exciting of sufficiently high order allows Willems' fundamental
lemma to parameterize every admissible trajectory of length
$T_{\mathrm{ini}}+N$ using the columns of the partitioned Hankel matrices~\cite{willems2005fundamental}.

At each control step $k$, DeePC embeds the trajectory representation directly into a finite-horizon optimal control problem. We define the predicted input and output sequences as
\begin{equation}
\vect u_k
=
\begin{bmatrix}
u_{k|k}\\
\vdots\\
u_{k+N-1|k}
\end{bmatrix},
\qquad
\vect y_k
=
\begin{bmatrix}
y_{k|k}\\
\vdots\\
y_{k+N-1|k}
\end{bmatrix}.
\label{eq:deepc_sequences}
\end{equation}
where $u_{k+i|k}$ and $y_{k+i|k}$ denote the input and output predicted
$i$ steps ahead using information available at control step $k$.

The initial trajectories comprise the $T_{\mathrm{ini}}$ most recently
applied inputs and their corresponding measured outputs:
$\vect u_{\mathrm{ini},k}
=[u_{k-T_{\mathrm{ini}}},\ldots,u_{k-1}]^{\mathsf T}$ and
$\vect y_{\mathrm{ini},k}
=[y_{k-T_{\mathrm{ini}}},\ldots,y_{k-1}]^{\mathsf T}$.

In the deterministic noise-free setting, the standard DeePC problem can be
written as
\begin{equation}
\begin{aligned}
\min_{\vect g,\vect u_k,\vect y_k}\quad
& \norm{\vect y_k-\vect r_k}_{Q}^{2}
+\norm{\vect u_k}_{R}^{2} \\
\text{subject to}\quad & \\
&
\begin{bmatrix}
U_{\mathrm P}\\
Y_{\mathrm P}\\
U_{\mathrm F}\\
Y_{\mathrm F}
\end{bmatrix}
\vect g
=
\begin{bmatrix}
\vect u_{\mathrm{ini},k}\\
\vect y_{\mathrm{ini},k}\\
\vect u_k\\
\vect y_k
\end{bmatrix},\\
& \vect u_k\in\mathcal U,\qquad
\vect y_k\in\mathcal Y.
\end{aligned}
\label{eq:noise_free_deepc}
\end{equation}
where $\vect r_k$ is the reference trajectory, $Q\succeq0$ weighs the
output-tracking error, and $R\succ0$ weighs the control effort and $\vect g\in\mathbb{R}^{H_c}$ is a coefficient vector, whose entries combine the recorded trajectories and map them to the candidate future input--output trajectories..

Problem~\eqref{eq:noise_free_deepc} is solved in a receding-horizon fashion.
At control step $k$, after calculating the optimal input sequence $\vect u_k^\star$,
only its first component $u_{k|k}^\star$ is applied to the system.
After obtaining the next output measurement, the problem is reinitialized by
updating
$[\vect u_{\mathrm{ini},k}^{\mathsf T}\;
 \vect y_{\mathrm{ini},k}^{\mathsf T}]^{\mathsf T}$
with the most recent input--output measurements and advancing the control
step from $k$ to $k+1$. ~\eqref{eq:noise_free_deepc} is then solved
again to calculate a new optimal input sequence over the next $N$ steps.
For deterministic LTI systems, DeePC is equivalent to 
model predictive control (MPC)~\cite{morari1999mpc}.

\subsection{Robust and Kernelized DeePC (RoKDeePC)}

Standard DeePC relies on the behavioral representation of deterministic LTI
systems. RoKDeePC extends data-driven predictive control to nonlinear systems
by replacing the linear predictor with a regularized kernel-based
multistep predictor \cite{lian2021nonlinear,huang2024rokdeepc}. In particular,
the future outputs are represented as
\begin{equation}
y_{\mathrm{[i]}}
=
f_i\!\left(
\vect u_{\mathrm{ini}},
\vect y_{\mathrm{ini}},
\vect u
\right),
\qquad
i=1,\ldots,pN,
\label{eq:nonlinear_multistep_predictor}
\end{equation}
where each $f_i$ predicts one component of the output trajectory over the
$N$-step horizon.

It is assumed that the functions $f_i$ are continuous and belong to a reproducing kernel Hilbert space (RKHS)
$\mathcal H_{\kappa}$ associated with a kernel $\kappa$. Let
\begin{equation}
\vect{x}_j
=
\begin{bmatrix}
U_{\mathrm{P}}(:,j)\\
Y_{\mathrm{P}}(:,j)\\
U_{\mathrm{F}}(:,j)
\end{bmatrix},
\qquad j=1,\ldots,H_c.
\end{equation}
denote the $j$th recorded past-input, past-output, and future-input trajectory.
Each multistep predictor is obtained through regularized kernel regression:
\begin{equation}
\hat f_i
=
\min_{f_i\in\mathcal H_{\kappa}}
\sum_{j=1}^{H_c}
\left(
[Y_{\mathrm F}]_{ij}-f_i(\vect x_j)
\right)^2
+
\gamma\norm{f_i}_{\mathcal H_{\kappa}}^2
\label{eq:rkhs_regression}
\end{equation}
By the representer theorem~\cite{scholkopf2001generalized}, this minimization problem achieves a closed-form solution:
\begin{equation}
{\vect y}
=
Y_{\mathrm F}
\left(K+\gamma I\right)^{-1}
\vect k(\vect z),
\label{eq:kernel_predictor}
\end{equation}
where
$\vect z=
\begin{bmatrix}
\vect u_{\mathrm{ini}}^{\mathsf T} &
\vect y_{\mathrm{ini}}^{\mathsf T} &
\vect u^{\mathsf T}
\end{bmatrix}^{\mathsf T}$, $K\in\mathbb{S}_{\succeq 0}^{H_c\times H_c}$ is real symmetric positive-definite matrix also known as the Gram matrix, $[K]_{ij}=\kappa(\vect x_i,\vect x_j)$, and $\vect k(\vect z)=
\begin{bmatrix}
\kappa(\vect x_1,\vect z) &
\cdots &
\kappa(\vect x_{H_c},\vect z)
\end{bmatrix}^{\mathsf T}$, $\gamma>0$ regularizes the predictor.

A kernel can be chosen depending on prior knowledge of the system. We use a Gaussian Kernel $\kappa(\vect x_i,\vect x_j)=\exp\left(-\norm{\vect x_i-\vect x_j}^2/(2\sigma^2)\right)$ which provides a smooth similarity measure for representing nonlinear relationships between the input--output trajectories.

Defining $K_{\gamma}=K+\gamma I$, the predictor can equivalently be written
in the implicit form
\begin{equation}
K_{\gamma}\vect g=\vect k(\vect z),
\qquad
\vect y=Y_{\mathrm F}\vect g.
\label{eq:implicit_kernel_predictor}
\end{equation}

The optimization problem with the non-linear multi step predictor can be written as:
\begin{equation}
\begin{aligned}
\min_{\substack{
\vect u\in\mathcal U,\,
\vect y\in\mathcal Y,\,
\vect g}}
\quad
& \ell(\vect u)+\norm{\vect y-\vect r}_{Q}^{2}\\
\text{subject to}\quad
& \eqref{eq:implicit_kernel_predictor}.
\end{aligned}
\end{equation}
where $\ell(\vect u)$ penalizes the control input, $\mathcal U$ and
$\mathcal Y$ specify the sets of input and output
values available for selection by the optimizer, respectively.
The constraints in~\eqref{eq:implicit_kernel_predictor} are moved to the cost function as penalties to handle inexact data and simplify solver design.  Including the robustification of~\eqref{eq:implicit_kernel_predictor}, the new problem can be framed as
\begin{equation}
\begin{aligned}
\min_{\vect u\in\mathcal U,\,\vect g\in\mathcal G}\quad
    &\ell(\vect u)
     +\norm{Y_{\mathrm F}\vect g-\vect r}_{Q}^{2}
     +\lambda_g\norm{\vect g}^{2}\\
    &+\lambda_k
     \norm{
     K_{\gamma}\vect g-\vect k(\vect z)
     }^{2}
\end{aligned}
\label{eq:quadratic_rokdeepc}
\end{equation}
where $\lambda_k$ weighs the soft constraint and can be understood as a weight on kernel consistency. $\lambda_g$ comes from the robust treatment to the optimization problem and can be understood as a weight regularizing the trajectory coefficients. 

We refer the reader to~\cite{huang2024rokdeepc} for further discussion on the robustification and performance guarantees under disturbance.

Problem~\eqref{eq:quadratic_rokdeepc} is nonconvex in $\vect u$ due to the nonlinear dependence of the Gaussian kernel $\vect k(\vect z)$ on $\vect u$. However, for fixed $\vect u$, it is convex in $\vect g$, provided that $Q\succeq0$, $\lambda_g,\lambda_k\geq0$, and $\mathcal G$ is convex.

When the output constraints $\vect g\in\mathcal G$ are inactive in~\eqref{eq:quadratic_rokdeepc}, the minimizer with fixed $\vect u$ has the closed-form solution
\begin{equation}
\begin{aligned}
\vect g^\star(\vect u)
&=
M_1^{-1}
\left(
Y_{\mathrm F}^{\mathsf T}Q\vect r
+\lambda_kK_{\gamma}
\vect k(\vect z)
\right)\\
&=
M_r\vect r
+
M_k\vect k(\vect z),
\end{aligned}
\label{eq:gstar}
\end{equation}
where $M_1=Y_{\mathrm F}^{\mathsf T}QY_{\mathrm F}+\lambda_g I+\lambda_kK_{\gamma}^{2}$, $M_r=M_1^{-1}Y_{\mathrm F}^{\mathsf T}Q$, and $M_k=M_1^{-1}\lambda_kK_{\gamma}$.

RoKDeePC substitutes the closed-form solution~\eqref{eq:gstar} into
Problem~\eqref{eq:quadratic_rokdeepc}, reducing the optimization to the
selection of the future input sequence $\vect u$. The authors
of~\cite{huang2024rokdeepc} solve the resulting nonconvex problem using a
projected-gradient descent method. In their power-converter example, solving the full nonlinear RoKDeePC problem required approximately $0.6\,\mathrm{s}$ per
control update, substantially exceeding the system sampling period of
$2\,\mathrm{ms}$. To reduce this computation time, they considered a kernel
that remains nonlinear in the past input--output data but is globally linear
in the future input sequence. This assumption reduced the control problem to a quadratic program and decreased the reported computation time to
approximately $10\,\mathrm{ms}$. However, robotic systems commonly exhibit
configuration-dependent dynamics and nonlinear actuation for which future outputs may depend nonlinearly on the applied inputs~\cite{murray1994mathematical,hogan2022contact}. Our proposed method retains the nonlinear Gaussian kernel, replaces the projected-gradient descent method with damped Gauss--Newton steps, and uses a hybrid buffer that incorporates informative online data. The following section describes our method.

\section{METHOD}
\label{Section: Method}



The section consists of two components: a kernelized
data-enabled predictor with Gauss--Newton update steps for fast, real-time control, and a hybrid buffer that adapts the predictor when the observed dynamics change.

\subsection{Kernelized Data Predictor with Gauss--Newton updates}

We use the quadratic RoKDeePC problem~\eqref{eq:quadratic_rokdeepc} and its
closed-form inner solution~\eqref{eq:gstar}, introduced in
Section~\ref{Sec:Background}. The authors of~\cite{huang2024rokdeepc}
address the nonconvex dependence on $\vect u$ using a projected-gradient
method. For real-time implementation, they additionally restrict the kernel
to be globally linear in the future input sequence, thereby reducing the
optimization to a quadratic program. We do not impose this restriction;
instead, we retain the nonlinear Gaussian kernel and exploit the nonlinear
least-squares structure of the reduced RoKDeePC objective using a damped, Gauss--Newton method.

Substituting $\vect g^*(\vect u)$ from~\eqref{eq:gstar} in~\eqref{eq:quadratic_rokdeepc}, the new optimization problem can be written as:
\begin{equation}
\begin{aligned}
\min_{\vect u\in\mathcal U}\quad
&
\ell(\vect u)
+
\left\|
Y_{\mathrm F}
\left(
M_r\vect r+M_k\vect k(\vect u)
\right)
-\vect r
\right\|_Q^2
\\
&+
\lambda_g
\left\|
M_r\vect r+M_k\vect k(\vect u)
\right\|^2
\\
&+
\lambda_k
\left\|
K_\gamma
\left(
M_r\vect r+M_k\vect k(\vect u)
\right)
-\vect k(\vect u)
\right\|^2 
\label{eq:reduced_rokdeepc}
\end{aligned}
\end{equation}

where $M_1$, $M_r$, and $M_k$ are as in~\eqref{eq:gstar}.

We further define the input cost as: 
\begin{equation}
\ell(\vect u)
=
W_U\|\vect u\|^2
+
W_{\Delta U}\|\Delta\vect u\|^2,
\label{eq:input_cost}
\end{equation}
where $\Delta\vect u$ denotes the difference between consecutive control
inputs, i.e., $\Delta u_i=u_i-u_{i-1}$. $W_U$ penalizes large force commands and the $W_{\Delta U}$ penalizes the rate of change in force commands between consecutive control steps.

The problem~\eqref{eq:reduced_rokdeepc} is a nonlinear least squares problem. It can be written as:
\begin{equation}
    \min_{\vect u}
    \left\lVert
        \vect\rho(\vect u)
    \right\rVert^2,
    \label{eq:nonlinear_least_squares}
\end{equation}
where the residual vector is
\begin{equation}
\vect\rho(\vect u)
=
\begin{bmatrix}
Q^{1/2}
\left[
Y_{\mathrm F}
\left(
M_r\vect r+M_k\vect k(\vect z)
\right)
-\vect r
\right]
\\[2mm]
\sqrt{\lambda_g}
\left(
M_r\vect r+M_k\vect k(\vect z)
\right)
\\[2mm]
\sqrt{\lambda_k}
\left[
K_{\gamma}
\left(
M_r\vect r+M_k\vect k(\vect z)
\right)
-\vect k(\vect z)
\right]
\\[2mm]
\sqrt{W_U}\,\vect u
\\[1mm]
\sqrt{W_{\Delta U}}\,\Delta\vect u
\end{bmatrix}.
\label{eq:gn_residual}
\end{equation}

At Gauss--Newton iteration $i$, the Gaussian kernel vector is approximated
locally using its first-order expansion:
\begin{equation}
    \vect k(\vect u^{(i)}+\Delta\vect u)
    \approx
    \vect k(\vect u^{(i)})
    +
    J_k(\vect u^{(i)})\Delta\vect u,
    \label{eq:kernel_linearization}
\end{equation}
where $J_k(\vect u)=\partial\vect k(\vect u)/\partial\vect u$ is the analytical Jacobian of the Gaussian kernel vector. Equation~\eqref{eq:kernel_linearization} provides a local linear approximation of the otherwise nonlinear kernel vector. After updating the control sequence, the exact nonlinear kernel vector and the corresponding objective value are evaluated again.

Because $\partial\vect g^*/\partial\vect u=M_kJ_k$, the residual Jacobian is
\begin{equation}
J_{\rho}
=
\begin{bmatrix}
Q^{1/2}Y_{\mathrm F}M_kJ_k\\
\sqrt{\lambda_g}\,M_kJ_k\\
\sqrt{\lambda_k}(K_{\gamma}M_k-I)J_k\\
\sqrt{W_U}\,I\\
\sqrt{W_{\Delta U}}\,J_{\Delta u}
\end{bmatrix},
\label{eq:gn_jacobian}
\end{equation}
where $J_{\rho}=\partial\vect \rho/\partial\vect u$ and $J_{\Delta u}=\partial(\Delta\vect u)/\partial\vect u$.

Using this Jacobian, the nonlinear residual vector is approximated locally as
\begin{equation}
\vect{\rho}\!\left(\vect{u}^{(i)}+\delta\vect{u}\right)
\approx
\vect{\rho}^{(i)}
+
J_{\rho}^{(i)}\delta\vect{u},
\label{eq:residual_linearization}
\end{equation}
where $J_{\rho}^{(i)}=J_{\rho}(\vect{u}^{(i)})$ and
$\vect{\rho}^{(i)}=\vect{\rho}(\vect{u}^{(i)})$ denote the residual Jacobian
and residual vector evaluated at iteration $i$.

The damped Gauss--Newton search direction is obtained by solving the local
least-squares problem
\begin{equation}
\min_{\delta\vect{u}}
\left\|
\vect{\rho}^{(i)}
+
J_{\rho}^{(i)}\delta\vect{u}
\right\|^2
+
\eta\|\delta\vect{u}\|^2.
\label{eq:local_gn_problem}
\end{equation}
where $\eta>0$ is the damping coefficient.
Its first-order optimality condition gives
\begin{equation}
\left[
\left(J_{\rho}^{(i)}\right)^{\mathsf T}J_{\rho}^{(i)}
+\eta I
\right]
\delta\vect{u}^{(i)}
=
-\left(J_{\rho}^{(i)}\right)^{\mathsf T}\vect{\rho}^{(i)}.
\label{eq:gn_step}
\end{equation}

which can be solved efficiently.

The candidate force sequence is then
updated according to
\begin{equation}
\vect u^{(i+1)}
=
\operatorname{proj}_{\vect u\in\mathcal U}
\left(
\vect u^{(i)}+\delta\vect u
\right)
\label{eq:projected_gn_update}
\end{equation}

The projection ensures that every candidate input remains within the set $\mathcal U$. If a candidate input increases the cost of the original nonlinear objective in~\eqref{eq:reduced_rokdeepc}, the step
is rejected and recomputed with a larger damping coefficient.

The optimized input sequence from the previous step is shifted by one element and used to initialize the next optimization in a receding-horizon manner. Only the first optimized input is applied at each control step. The input that has already been applied is discarded, and the terminal value is repeated to preserve the horizon length:
\begin{equation}
\vect{u}_k^{(0)}
=
\begin{bmatrix}
u_{k|k-1}^{\star} &
\cdots &
u_{k+N-2|k-1}^{\star} &
u_{k+N-2|k-1}^{\star}
\end{bmatrix}^{\mathsf T}.
\label{eq:shifted_warm_start}
\end{equation}
where $\vect{u}_k^{(0)}$ is the initial guess at the control step $k$. 

\begin{algorithm}[!t]
\caption{RoKON}
\label{alg:adaptive_rokdeepc}
\begin{algorithmic}[1]

\Require Initial buffer $\mathcal B_0$ and feasible sequence $\vect{u}^{0}$

\State $\mathcal B\gets\mathcal B_0$ and rebuild the predictor matrices
\State $\vect{u}^{(0)}\gets\vect{u}^{0}$,
       $n_{\mathrm{pending}}\gets0$

\For{$k=0,\ldots,N_{\mathrm{step}}-1$}

    \If{a background reconstruction has finished}
        \State Install all reconstructed matrices
    \EndIf

    \If{$k>0$}
        \State Warm-start $\vect{u}^{(0)}$ using
        \eqref{eq:shifted_warm_start}
    \EndIf

    \State $\vect{u}_{k}^{\star}
    \gets
    \Call{DampedGN}{\vect{u}^{(0)}}$
    using~\eqref{eq:gn_residual}--\eqref{eq:projected_gn_update}

    \State $\hat y_{k+1}^{-}
    \gets
    [Y_{\mathrm F}\vect{g}^{\star}
    (\vect{u}_{k}^{\star})]_1$

    \State Apply $u_k\gets[\vect{u}_{k}^{\star}]_1$
    and measure $y_{k+1}$

    \State Update $\vect{u}_{\mathrm{ini}}$ and
    $\vect{y}_{\mathrm{ini}}$

    \State $e_{k+1}
    \gets y_{k+1}-\hat y_{k+1}^{-}$

    \If{a valid contiguous column $\mathcal C_{k+1}$ is available}

        \If{$|\mathcal B|<H_{\max}$}
            \State Append $\mathcal C_{k+1}$ to
            $\mathcal B^{\mathrm{r}}$;
            $n_{\mathrm{pending}}\gets n_{\mathrm{pending}}+1$

        \ElsIf{$|e_{k+1}|
        >\varepsilon_{\mathrm{online}}$}
            \State Replace the oldest replaceable column by
            $\mathcal C_{k+1}$;
            $n_{\mathrm{pending}}\gets n_{\mathrm{pending}}+1$
        \EndIf

    \EndIf

    \If{$n_{\mathrm{pending}}\geq n_{\mathrm{batch}}$
    and no reconstruction is active}
        \State Launch
        $\Call{Rebuild}{\operatorname{copy}(\mathcal B)}$
        in the background
        \State $n_{\mathrm{pending}}\gets0$
    \EndIf

\EndFor

\end{algorithmic}
\end{algorithm}

\subsection{Hybrid Buffer}
\label{Sec:HybridBuffer}

The predictor is initialized from a time-ordered offline sequence of measured
input--output pairs. Using the construction in
\eqref{eq:hankel_partition}, this sequence is converted into overlapping,
contiguous Hankel-matrix columns of length
$L=T_{\mathrm{ini}}+N$. Each buffer entry therefore contains the past and
future components $(U_{\mathrm{P}(:,j)},Y_{\mathrm P(:,j)},U_{\mathrm F(:,j)}, Y_{\mathrm F(:,j)})$ associated with one complete trajectory window. Forming
each entry from contiguous samples prevents artificial transitions from being
introduced within a trajectory column.

Let $H_{\max}$ denote the maximum number of Hankel-matrix columns used to construct the buffer and let $H_{\mathrm p}$
denote the number of protected offline columns. The buffer at control step
$k$ is partitioned as
\begin{equation}
\mathcal B_k
=
\mathcal B^{\mathrm{p}}
\cup
\mathcal B_k^{\mathrm{r}},
\qquad
|\mathcal B_k|\leq H_{\max}.
\label{eq:buffer_partition}
\end{equation}
where $\mathcal B_k$ denotes the complete buffer available to the predictor at
control step $k$, $\mathcal B^{\mathrm p}$ contains the protected offline
Hankel-matrix columns that remain fixed throughout the experiment, and
$\mathcal B_k^{\mathrm r}$ contains the replaceable Hankel-matrix columns that may be updated using online measurements. The first $H_{\mathrm p}$ offline columns are assigned to
$\mathcal B^{\mathrm{p}}$ and remain unchanged throughout the
closed-loop run. These columns preserve a fixed set of previously observed
dynamics and prevent complete forgetting during online adaptation. All
remaining offline and online columns belong to
$\mathcal B_k^{\mathrm{r}}$ and may be removed as new measurements
become available.

Online admission is based on the one-step-ahead prediction residual. Let
$\hat y_k^{-}$ denote the velocity predicted by the active model before the
measurement $y_k$ is available. The signed residual is
\begin{equation}
e_k
=
y_k-\hat y_k^{-}.
\label{eq:signed_prediction_residual}
\end{equation}
This residual is independent of the output reference and measures the mismatch between the predicted output and the observed response to the applied input.

A newly completed trajectory column is considered informative when
\begin{equation}
|e_k|
>
\varepsilon_{\mathrm{online}},
\label{eq:buffer_admission_condition}
\end{equation}
where $\varepsilon_{\mathrm{online}}$ is the admission threshold. Before the buffer reaches
$H_{\max}$, informative online columns are appended without removing existing data.
After the buffer is full, an admitted column replaces the oldest entry in
$\mathcal B_k^{\mathrm{r}}$. Columns in
$\mathcal B^{\mathrm{p}}$ are never selected for replacement.

Changing the buffer composition changes the matrices
$U_{\mathrm P}$, $Y_{\mathrm P}$, $U_{\mathrm F}$, and $Y_{\mathrm F}$ in
\eqref{eq:hankel_partition}. It also changes the kernel Gram matrix
$K$, and the matrices $M_r$ and $M_k$
appearing in the closed-form solution~\eqref{eq:gstar}. These quantities must therefore be reconstructed whenever an updated buffer
is installed. Further details on the installation of buffer are provided in~\ref{Implementation}.

The complete algorithm for our proposed method is defined in~\eqref{alg:adaptive_rokdeepc}. The $\Call{DampedGN}{\cdot}$ performs at most
$N_{\mathrm{GN}}$ Gauss--Newton iterations. The $\Call{Rebuild}{\cdot}$ reconstructs the trajectory, kernel, and
solver matrices from an immutable copy of the current buffer.

\section{Experimental Evaluation}
\label{Section: Experiments}

This section presents the experimental setup and implementation details,
describes the experiments conducted, and discusses the results.

\subsection{Setup}

The MuJoCo scene comprises a Franka Research~3 (FR3) arm, a rigid block
with mass $0.5\,\mathrm{kg}$ and dimensions
$0.10\times0.10\times0.05\,\mathrm{m}$, and three surfaces with friction coefficients $\mu\in\{0.2,\,0.5,\,0.25\}$ kept on a table.
The arm's gripper pushes the block in a single direction across all three
surfaces.

At control step $k$, let $u_k\in\mathbb{R}$ denote the force applied by the end effector and let $y_k\in\mathbb{R}$ denote the block velocity along the pushing
axis. The sampled input--output dynamics are represented abstractly as
\begin{equation}
\vect y_k
=
\mathcal F_k\!\left(
\vect u_{\mathrm{ini},k},
\vect y_{\mathrm{ini},k},
\vect u_k
\right)
+
\vect w_k,
\label{eq:unknown_dynamics}
\end{equation}
where $\mathcal F_k$ is unknown and may change as the object crosses varying friction surfaces, and $w_k$ represents noise and unmodeled contact effects.

The control objective is to track a constant velocity reference
$y^{\mathrm{ref}}$ across surfaces with varying friction, without an explicit dynamics or friction model, using only measured force--velocity input--output data.

The controller is required to:
\begin{enumerate}
    \item initialize from a compact offline dataset collected only on the
    starting surface;
    \item operate without knowledge of the friction coefficient or the
    location of a surface transition;
    \item adapt its data-driven predictor using measurements collected during
    the push;
    \item generate control decisions sufficiently fast for real-time implementation.
\end{enumerate}

We conduct two sets of experiments. The first set evaluates velocity
tracking on uniform-friction surfaces, while the second evaluates velocity tracking on varying-friction surfaces using the proposed hybrid buffer. All experiments use a sampling period of $T_s=0.03\,\mathrm{s}$, a past horizon of $T_{\mathrm{ini}}=2$, a prediction horizon of $N=5$, and force range of $0$--$8\,\mathrm{N}$.

\subsubsection{Uniform-Friction Experiments}

These experiments investigate whether the controller can track a
constant velocity reference when the friction coefficient remains unchanged
throughout the run. Tests are conducted for $\mu\in\{0.08,\,0.25,\,0.50\}$ with
$y^{\mathrm{ref}}\in\{0.05,\,0.10\}\,\mathrm{m/s}$.

An offline input--output trajectory $\{(u_k^{\mathrm d},y_k^{\mathrm d})\}_{k=0}^{T-1}$ is collected by
operating the arm with piecewise-constant force commands. The force is varied in increments of $0.5\,\mathrm{N}$ over the range
$[0,8]\,\mathrm{N}$ to collect sufficiently diverse force--velocity pairs
for constructing the offline buffer. Online adaptation is disabled for these experiments because each trial is conducted on a single surface with a constant friction coefficient.

\subsubsection{Varying-Friction Experiment}
\label{Varying-Friction}

The second experiment evaluates closed-loop tracking and online adaptation while the block crosses three surfaces with different friction coefficients. The block traverses the three surfaces in the order    $\mu=0.20\;\rightarrow\; \mu=0.50 \;\rightarrow\; \mu=0.25,$
where each surface has a length of $0.40\,\mathrm{m}$. The reference velocity is $y^{\mathrm{ref}}=0.02\,\mathrm{m/s}.$
The controller is initialized using an
offline buffer containing data collected only on the first surface
($\mu=0.20$). It is not provided with the friction coefficients of the subsequent
surfaces or the locations of the surface transitions on the table.

We compare our method with four controllers. Stribeck MPC uses an explicit nonlinear friction model~\cite{canudas1995friction} with a fixed friction coefficient $\mu=0.2$. Linear DeePC~\cite{coulson2019deepc} uses a fixed $294$-column Hankel matrix constructed from the offline trajectory.
RoKDeePC~\cite{huang2024rokdeepc} uses the same fixed offline buffer with the projected-gradient method to solve the non-linear objective. Windowed DeePC follows the batch-reconstruction principle in~\cite{leng2026deepconverter}, periodically reconstructing its Hankel
matrix from a recent contiguous trajectory window.

\subsection{Implementation}
\label{Implementation}

The parameters are manually tuned and set to
$Q=8\times10^{3}I_N$, $\lambda_k=5\times10^{6}$,
$\lambda_g=1$, $W_U=10^{-4}$, and $W_{\Delta U}=3$. The Gaussian-kernel parameters are manually tuned and fixed to $\sigma^2=4$ and
$\gamma=0.02$.

For ~\eqref{Varying-Friction} the initial offline trajectory contains $T=300$ samples. With
$T_{\mathrm{ini}}=2$ and $N=5$, this produces $H_c=T-T_{\mathrm{ini}}-N+1=294$
overlapping trajectory columns. These columns initialize the hybrid buffer.
The total buffer capacity and number of protected columns are $H_{\max}=350, H_{\mathrm p}=100.$ The oldest replaceable column is replaced only when the one-step prediction residual satisfies \eqref{eq:buffer_admission_condition}, with $\varepsilon_{\mathrm{online}}
    =10^{-3}\,\mathrm{m/s}.$
    
Reconstructing the predictor after every admitted column would introduce
unnecessary and irregular computation for real-time implementation. Therefore, newly admitted Hankel matrix columns are accumulated into a small batch. Once a fixed number of columns get accumulated, the current buffer is copied into an immutable snapshot and passed to a non-real-time background worker on the CPU. The worker reconstructs the trajectory matrices, evaluates the kernel Gram matrix, forms the matrix used in the closed-form solution for $\vect{g}^{\star}$, and calculates $M_r$ and $M_k$ using optimized BLAS/LAPACK matrix operations and linear solves.

The previous complete buffer and matrices continue to be used while this reconstruction is in progress. After the worker finishes, the newly constructed matrices are installed together through an atomic model swap at an update boundary. Consequently, one control decision cannot combine trajectory matrices from one buffer version with kernel or solver matrices from another. The buffer capacity bounds the reconstruction cost, while batching and background execution prevent the dense matrix operations from delaying the online optimization or other time-critical control computations. The number of BLAS worker threads is also limited to reduce computational interference during real-time execution.

\subsubsection*{Low-Level Controller}

A desired Cartesian pushing force cannot be commanded directly to the arm's
joints. The desired force must first be represented as a Cartesian wrench and then mapped to joint torques. The low-level arm controller operation at $1\,\mathrm{kHz}$ performs this mapping while also maintaining the
end-effector position and orientation in the directions that do not contribute to pushing.

   \begin{figure}
      \centering
      \includegraphics[width=1\linewidth]{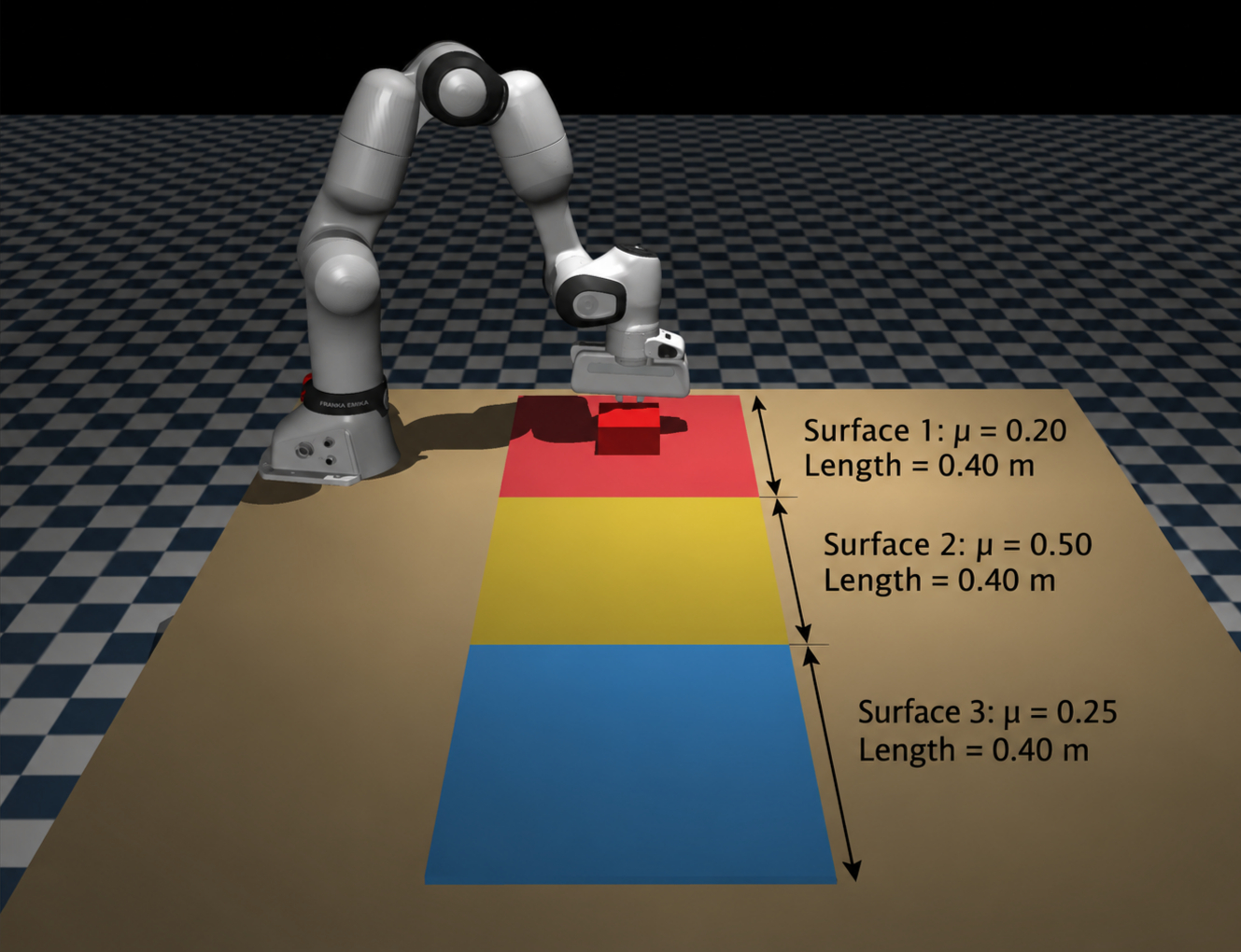}
      \caption{Image of the setup on MuJoCo. An FR3 arm pushing a block across three variable friction surfaces.}
      \label{mujoco_setup}
   \end{figure}



\subsection{Results}
\label{Sec: Results}

Tracking performance across the two experiments is reported using the root-mean-square velocity error. For a set of samples
$\mathcal I_s$ belonging to surface $s$, the surface RMSE is $
    \sqrt{
        \frac{1}{|\mathcal I_s|}
        \sum_{k\in\mathcal I_s}
        \left(y_k-y^{\mathrm{ref}}\right)^2
    }$.

\subsubsection{Uniform-Friction Experiments}

\begin{figure}
      \centering
      \includegraphics[width=1\linewidth,trim={0 1.5cm 0 0},clip]{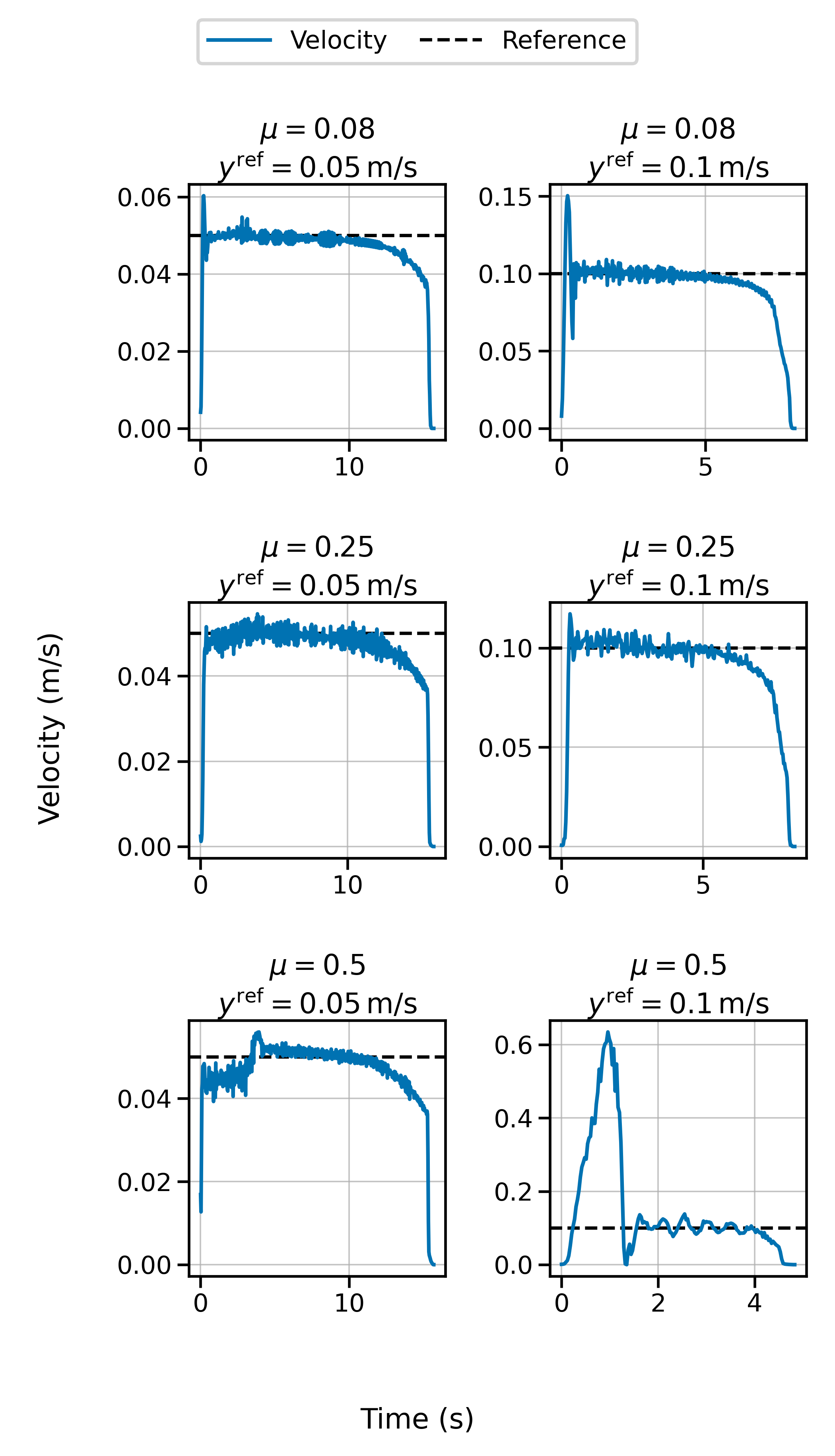}
      \includegraphics[width=1\linewidth, trim={0 0 0 14.7cm},clip]{single_surface.png}
      \caption{Closed-loop velocity tracking performance on uniform-friction surfaces for $\mu\in\{0.08,\,0.25,\,0.50\}$ and $y^{\mathrm{ref}}\in\{0.05,\,0.10\}\,\mathrm{m/s}$.}
      \label{single_surface_experiments}
   \end{figure}

Figure~\ref{single_surface_experiments} presents the closed-loop velocity tracking for
$\mu\in\{0.08,\,0.25,\,0.50\}$ and
$y^{\mathrm{ref}}\in\{0.05,\,0.10\}\,\mathrm{m/s}$.

The loss of tracking near the end of each experiment is caused by the arm
approaching the boundary of its reachable workspace. At this point, the end
effector can no longer continue moving in the pushing direction and therefore cannot execute the force requested by the control loop. 

\subsubsection{Varying-Friction Experiment}

Table~\ref{tab:surface_rmse_timing} reports the RMSE on each surface, the
complete-run RMSE, the number of Hankel columns $H_c$ used during closed-loop, and the mean control step computation time. A dash indicates that the block did not reach the corresponding surface. For controllers with a changing buffer, the minimum and maximum observed values of $H_c$ are reported.

Stribeck MPC and RoKDeePC track the initial surface accurately but degrade
as the friction changes and do not reach the third surface. Linear DeePC
completes the experiment with moderate tracking error. The proposed method achieves the lowest complete-run RMSE and the best tracking performance on the second and third surfaces. Linear DeePC is computationally faster than our method because its linear predictor leads to a convex quadratic program and avoids repeated kernel evaluations, Jacobian calculations, and nonlinear solver iterations.

\begin{table*}[t]
\caption{Varying Friction Experiment Tracking Accuracy and Computation Time}
\label{tab:surface_rmse_timing}
\centering
\begin{tabular}{|l||c|c|c||c||c||c|}
\hline
Method
& \multicolumn{3}{c||}{Surface RMSE (m/s)}
& Complete-run RMSE
& $H_c$ columns
& Mean time (ms) \\
\cline{2-4}
& $\mu=0.2$
& $\mu=0.5$
& $\mu=0.25$
& (m/s)
& 
& \\
\hline

Stribeck MPC
& 0.00370
& 0.01909
& --
& 0.01766
& --
& 7.20 \\

Linear DeePC
& 0.00709
& 0.00721
& 0.02502
& 0.01331
& 294
& \textbf{2.76} \\

RoKDeePC
& \textbf{0.00113}
& 0.01703
& --
& 0.01552
& 294
& 14.44 \\

Windowed DeePC
& 0.01496
& 0.00490
& 0.03687
& 0.02029
& 294--350
& 4.37 \\

RoKON (Ours)
& 0.00202
& \textbf{0.00308}
& \textbf{0.01084}
& \textbf{0.00602}
& 294--350
& 3.81 \\
\hline
\end{tabular}
\end{table*}

   \begin{figure}
      \centering
      \includegraphics[width=1\linewidth]{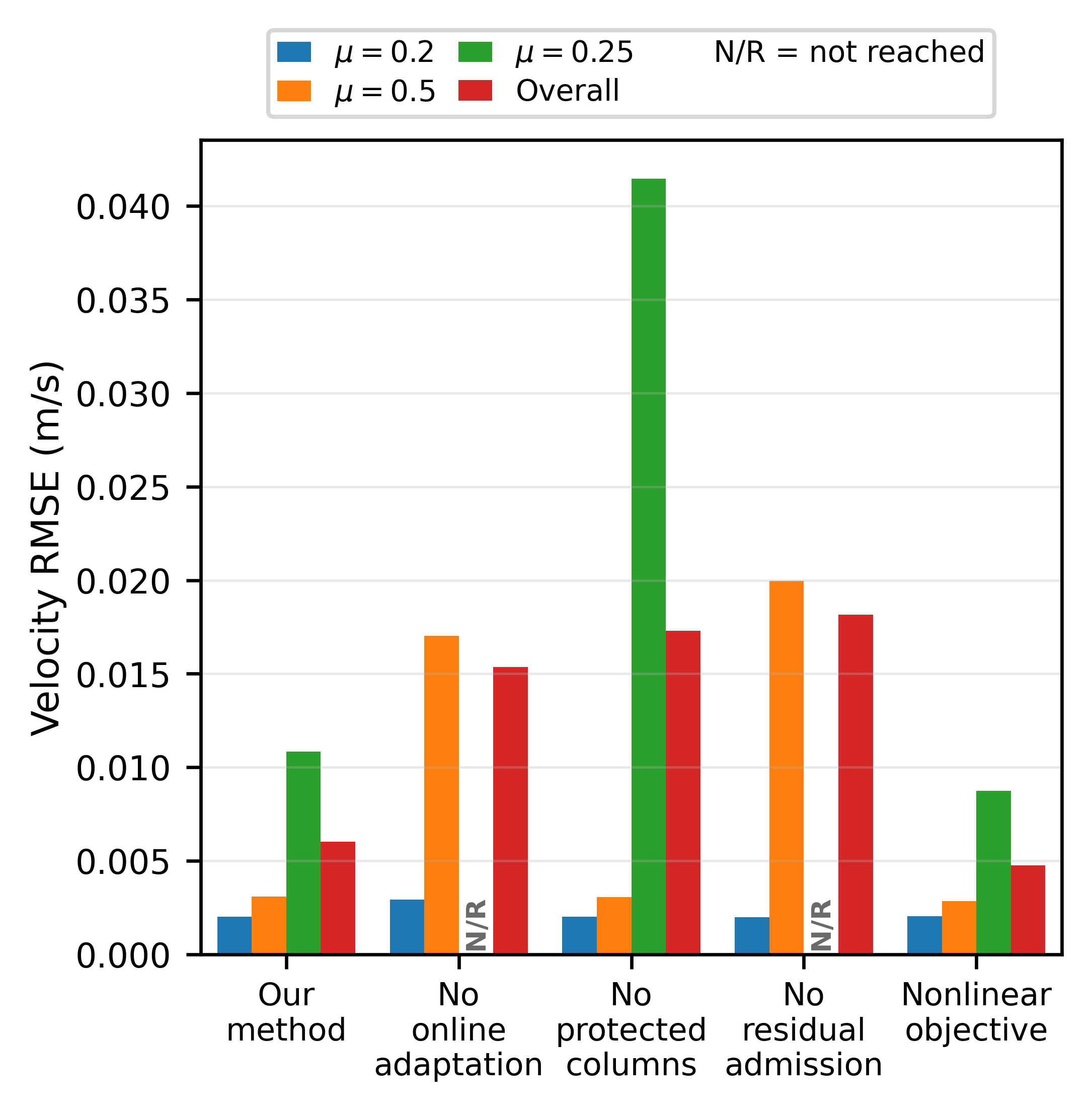}
      \caption{Ablation study of the proposed RoKON method across the three friction surfaces. The bars report the velocity-tracking RMSE on each surface and over the complete evaluated trajectory. N/R indicates that the block did not reach the corresponding surface.}
      \label{ablation}
   \end{figure}

\subsubsection{Ablation Study}

An ablation study is conducted to identify the contribution of the individual
components of our method. Four variants are compared with the complete method. In the \emph{no online
adaptation} variant, the initial offline buffer and predictor remain fixed
throughout the experiment. In the \emph{no protected columns} variant,
residual-based admission remains active, but every buffer column, including
the initial offline columns, may be replaced. In the \emph{no residual
admission} variant, the protected columns remain fixed, but every valid
completed trajectory is admitted without applying
\eqref{eq:buffer_admission_condition}. Finally, in the \emph{nonlinear
objective} variant, the damped Gauss--Newton method is replaced by direct
L-BFGS-B optimization of the RoKDeePC objective, while all other
adaptation components remain unchanged.

Figure~\ref{ablation} reports the surface and complete-run RMSE for each
variant. Without online adaptation, the method tracks the initial surface with an RMSE of approximately $0.0029\,\mathrm{m/s}$. After entering the second surface, however, its velocity remains substantially below the reference, increasing the second-surface RMSE to approximately $0.0170\,\mathrm{m/s}$. The block does not reach the third surface.

Removing the protected columns preserves good tracking on the first two
surfaces, but increases the third-surface RMSE to approximately
$0.0415\,\mathrm{m/s}$. Since every offline column is eligible for replacement, it cannot adapt to the third surface because previously observed behaviors may be progressively removed from the buffer.

Without residual-based admission, valid trajectories are admitted regardless
of whether they expose a meaningful prediction mismatch. The velocity
collapses on the second surface, and the block does not reach the third
surface. 

Replacing the Gauss--Newton solver with direct nonlinear optimization
produces the lowest RMSE among the evaluated variants over the completed
trajectory. However, it also has the highest mean control-decision time,
requiring $13.27\,\mathrm{ms}$ per update, compared with $3.72\,\mathrm{ms}$ for the proposed Gauss--Newton method.

\section{CONCLUSIONS}
\label{Sec: Conclusion}

This paper presented an adaptive kernelized data-enabled predictive control framework and showcased its performance on robotic pushing across surfaces with unknown and
changing friction. The proposed method builds on thenonlinear Gaussian-kernel predictor from prior work, but exploits the least-squares structure of
the control problem using damped Gauss--Newton updates, enabling control
inputs to be computed within the available sampling period. A hybrid buffer
combines protected offline data with informative online data
selected using the one-step prediction residual, allowing the predictor to
account for changes in the interaction dynamics while maintaining bounded
memory and computation.

Future work will evaluate the proposed method on the physical Franka
Research~3 arm and extend it to multidirectional manipulation tasks,
different objects, and more complex contact conditions. Incorporating
workspace, contact, and singularity constraints directly into the control
problem, together with automatic selection of the buffer
parameters, will also be investigated.

\addtolength{\textheight}{-12cm}   




\end{document}